\documentclass[10pt,twocolumn,letterpaper]{article}

\usepackage[pagenumbers]{cvpr} 

\usepackage{graphicx}
\usepackage{amsmath}
\usepackage{amssymb}
\usepackage{booktabs}
\usepackage{multirow}
\makeatletter
\@namedef{ver@everyshi.sty}{}
\makeatother
\usepackage{tikz}
\usetikzlibrary{positioning}
\usetikzlibrary{arrows}
\usetikzlibrary{shapes,snakes}
\usetikzlibrary{plotmarks}
\usepackage{pgfplots}
\pgfplotsset{compat=1.15}

\DeclareUnicodeCharacter{2212}{-}

\DeclareMathOperator*{\argmin}{arg\,min}

\usepackage[pagebackref,breaklinks,colorlinks]{hyperref}

\usepackage[capitalize]{cleveref}
\crefname{section}{Sec.}{Secs.}
\Crefname{section}{Section}{Sections}
\Crefname{table}{Table}{Tables}
\crefname{table}{Tab.}{Tabs.}

\def\cvprPaperID{8241} 
\def\confName{CVPR}
\def\confYear{2023}

\begin{document}

\title{Markerless Camera-to-Robot Pose Estimation via Self-supervised Sim-to-Real Transfer}

\author{Jingpei Lu, Florian Richter, and Michael C. Yip\\
University of California, San Diego\\
{\tt\small  \{jil360, frichter, yip\}@ucsd.edu}
}
\maketitle

\begin{abstract}
   Solving the camera-to-robot pose is a fundamental requirement for vision-based robot control, and is a process that takes considerable effort and cares to make accurate.
   Traditional approaches require modification of the robot via markers, and subsequent deep learning approaches enabled markerless feature extraction.
   Mainstream deep learning methods only use synthetic data and rely on Domain Randomization to fill the sim-to-real gap, because acquiring the 3D annotation is labor-intensive.
   In this work, we go beyond the limitation of 3D annotations for real-world data.
   We propose an end-to-end pose estimation framework that is capable of online camera-to-robot calibration and a self-supervised training method to scale the training to unlabeled real-world data.
   Our framework combines deep learning and geometric vision for solving the robot pose, and the pipeline is fully differentiable. To train the Camera-to-Robot Pose Estimation Network (CtRNet), we leverage foreground segmentation and differentiable rendering for image-level self-supervision. The pose prediction is visualized through a renderer and the image loss with the input image is back-propagated to train the neural network.
   Our experimental results on two public real datasets confirm the effectiveness of our approach over existing works. We also integrate our framework into a visual servoing system to demonstrate the promise of real-time precise robot pose estimation for automation tasks.
\end{abstract}

\section{Introduction}
\label{sec:intro}

The majority of modern robotic automation utilizes cameras for rich sensory information about the environment to infer tasks to be completed and provide feedback for closed-loop control.
The leading paradigm for converting the valuable environment information to the robot's frame of reference for manipulation is position-based visual servoing (PBVS)~\cite{chaumette2016visual}.
At a high level, PBVS converts 3D environmental information inferred from the visual data (e.g. the pose of an object to be grasped) and transforms it to the robot coordinate frame where all the robot geometry is known (e.g. kinematics) using the camera-to-robot pose.
Examples of robotic automation using the PBVS range from bin sorting \cite{mahler2019learning} to tissue manipulation in surgery \cite{li2020super}.

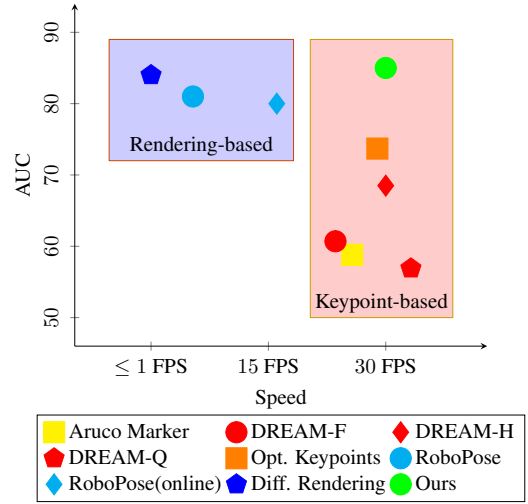
\begin{figure}[t!]
\centering

\begin{tikzpicture}[scale=0.8]
\begin{axis}[
legend style={
        at={(0.5,-0.2)},
        anchor=north,
        legend columns=3
    },
xlabel={Speed},
ylabel={AUC},
xtick={1.0,2.4,3.8},
xticklabels={$\leq 1$ FPS, $15$ FPS, $30$ FPS},
ymin=50,
ymax=90,
yticklabel style={rotate=90,anchor=base,yshift=0.2cm},
axis lines = left,
enlargelimits = true,
legend cell align={left},
]
\addplot[
scatter/classes={
ArucoMarker={mark=square*,yellow,mark size=5pt},%
DREAM-F={mark=oplus*,red,mark size=5pt},%
DREAM-H={mark=diamond*,red,mark size=5pt},%
DREAM-Q={mark=pentagon*,red,mark size=5pt},%
OptKey={mark=square*,orange,mark size=5pt},%
robopose={mark=oplus*,cyan,mark size=5pt},%
robopose-online={mark=diamond*,cyan,mark size=5pt},
diffrender={mark=pentagon*,blue,mark size=5pt},
Ours={mark=oplus*,green,mark size=5pt}
},
scatter,only marks,
scatter src=explicit symbolic,
]
table[x=speed,y=accuracy,meta=label]
{example.dat};
\legend{Aruco Marker,DREAM-F, DREAM-H,DREAM-Q, Opt. Keypoints, RoboPose, RoboPose(online), Diff. Rendering, Ours}
\addplot[patch,patch type=rectangle, fill opacity=0.2, blue]
coordinates {
(0.5,72) (0.5,89) (2.7,89) (2.7,72)
};
\node[] at (axis cs: 1.6,74) {Rendering-based};
\addplot[patch,patch type=rectangle, fill opacity=0.2, red]
coordinates {
(2.9,50) (2.9,89) (4.6,89) (4.6 , 50)
};
\node[] at (axis cs: 3.75,52) {Keypoint-based};
\end{axis}
\end{tikzpicture}
\caption{Comparison of speed and accuracy (based on AUC metric) for existing image-based robot pose estimation methods.}
\label{fig:methods_scatter} 
\end{figure}

Calibrating camera-to-robot pose typically requires a significant amount of care and effort. Traditionally, the camera-to-robot pose is calibrated with externally attached fiducial markers (\eg Aruco Marker~\cite{garrido2014aruco}, AprilTag~\cite{olson2011apriltag}). The 2D location of the marker can be extracted from the image and the corresponding 3D location on the robot can be calculated with forward kinematics. Given a set 2D-3D correspondence, the camera-to-robot pose can be solved using Perspective-n-Point (PnP) methods~\cite{lepetit2009epnp,gao2003p3p}.
The procedure usually requires multiple runs with different robot configurations and once calibrated, the robot base and the camera are assumed static. 
The incapability of online calibration limits the potential applications for vision-based robot control in the real world, where minor bumps or simply shifting due to repetitive use will cause calibrations to be thrown off, not to mention real-world environmental factors like vibration, humidity, and temperature, are non-constant. 
Having flexibility on the camera and robot is more desirable so that the robot can interact with an unstructured environment.

Deep learning, known as the current state-of-the-art approach for image feature extraction, brings promising ways for markerless camera-to-robot calibration.
Current approaches to robot pose estimation are mainly classified into two categories: keypoint-based methods ~\cite{lee2020dream,lu2022keypoint,richter2021robotic,lambrecht2019towards,lambrecht2021optimizing} and rendering-based methods~\cite{labbe2021robopose,hao2018vision}.
Keypoint-based methods are the most popular approach for pose estimation \textcolor{black}{because of the fast inference speed}.
However, the performance is limited to the accuracy of the keypoint detector which is often trained in simulation such that the proposed methods can generalize across different robotic designs.
Therefore, the performance is ultimately hampered by the sim-to-real gap, which is a long-standing challenge in computer vision and robotics~\cite{zhao2020sim}.

Rendering-based methods can achieve better performance by using the shape of the entire robot as observation, which provides dense correspondence for pose estimation.
The approaches in this category usually employ an iterative refinement process and require a reasonable initialization for the optimization loop to converge~\cite{li2018deepim}.
Due to the nature that iteratively render and compare is time- and energy-consuming, rendering-based methods are more suitable for offline estimation where the robot and camera are held stationary.
In more dynamic scenarios, such as a mobile robot, the slow computation time make the rendering-based methods impracticable to use.

In this work, we propose CtRNet, an end-to-end framework for robot pose estimation \textcolor{black}{which, at inference, uses keypoints for the fast inference speed and leverages the high performance of rendering-based methods for training to overcome the sim-to-real gap previous keypoint-based methods faced.}
Our framework contains a segmentation module to generate a binary mask of the robot and keypoint detection module which extracts point features for pose estimation. 
Since segmenting the robot from the background is a simpler task than estimating the robot pose and localizing point features on robot body parts, we leverage foreground segmentation to provide supervision for the pose estimation.
Toward this direction, we first pretrained the network on synthetic data, which should have acquired essential knowledge about segmenting the robot.
Then, a self-supervised training pipeline is proposed to transfer our model to the real world without manual labels. 
We connect the pose estimation to foreground segmentation with a differentiable renderer~\cite{kato2018neural,liu2019soft}. The renderer generates a robot silhouette image of the estimated pose and directly compares it to the segmentation result. Since the entire framework is differentiable, the parameters of the neural network can be optimized by back-propagating the image loss.

\textbf{Contributions}.
Our main contribution is the novel framework for image-based robot pose estimation together with a scalable self-training pipeline that utilizes unlimited real-world data to further improve the performance \textit{without \textcolor{black}{any manual} annotations.}
Since the keypoint detector is trained with image-level supervision, we effectively encompass the benefits from both keypoint-based and rendering-based methods, where previous methods were divided.
As illustrated in the \cref{fig:methods_scatter}, our method maintains high inference speed while matching the performance of the rendering-based methods.
Moreover, we integrate the CtRNet into a robotic system for PBVS and demonstrate the effectiveness on real-time robot pose estimation.

\section{Related Works}
\label{sec:related_works}

\subsection{Camera-to-Robot Pose Estimation}

The classical way to calibrate the camera-to-robot pose is to attach the fiducial markers~\cite{garrido2014aruco,olson2011apriltag} to known locations along the robot kinematic chain. The marker is detected in the image frame and their 3D position in the robot base frame can be calculated with forward kinematics. With the geometrical constraints, the robot pose can be then derived by solving an optimization problem~\cite{park_robot_1994,fassi2005hand,ilonen2011robust,horaud1995hand}.

Early works on markerless articulated pose tracking utilize a depth camera for 3D observation \cite{schmidt2014dart,pauwels2014real,michel2015pose,desingh2019factored}. For a high degree-of-freedom articulated robot, Bohg \etal~\cite{bohg2014robot} proposed a pose estimation method by first classifying the pixels in depth image to robot parts, and then a voting scheme is applied to estimate the robot pose relative to the camera. This method is further improved in~\cite{widmaier2016robot} by directly training a Random Forest to regress joint angles instead of part label.
However, these methods are not suitable for our scenario where only single RGB image is available.

More recently, as deep learning becomes popular in feature extraction, many works have been employing deep neural networks for robot pose estimation. Instead of using markers, a neural network is utilized for keypoint detection, and the robot pose is estimated through an optimizer (\eg PnP solver)~\cite{lambrecht2019towards,lee2020dream,lu2022keypoint,zuo2019craves}. To further improve the performance, the segmentation mask and edges are utilized to refine the robot pose~\cite{lambrecht2021optimizing,hao2018vision}.
Labb\'{e} \etal~\cite{labbe2021robopose} also introduces the \textit{render\&compare} method to estimate the robot pose by matching the robot shape.
These methods mainly rely on synthetically generated data for training and hope the network can generalize to the real world by increasing the variance in data generation.
Our method explicitly deals with the sim-to-real transfer by directly training on real-world data with self-supervision.


\subsection{Domain Adaptation for Sim-to-Real Transfer}

In computer vision and robotics, Domain Randomization (DR)~\cite{tobin2017domain} is the most widely used method for sim-to-real transfer due to its simplicity and effectiveness. The idea is to randomize some simulation parameters (e.g. camera position, lighting, background, etc.) and hope that the randomization captures the distribution of the real-world data. This technique has been applied to object detection and grasping ~\cite{tremblay2018training,borrego2018generic,hinterstoisser2018pre,tobin2018domain,horvath2022object}, and pose estimation~\cite{sundermeyer2018implicit, manhardt2018deep, tremblay2018deep,lambrecht2019towards,lee2020dream,lu2022keypoint,zuo2019craves, labbe2021robopose}. The randomization is usually tuned empirically hence it is not efficient.

Another popular technique for domain transfer is Domain Adaptation (DA), which is to find the feature spaces that share a similar distribution between the source and target domains~\cite{wang2018deep}. 
This technique has shown recent success in computer vision~\cite{hoffman2018cycada,chen2018domain,sankaranarayanan2018learning} and robotic applications~\cite{bousmalis2018using,james2019sim,gupta2017learning}.
In this work, instead of finding the latent space and modeling the distribution between the simulation and the real world, we perform sim-to-real transfer by directly training on the real-world data via a self-training pipeline.

\begin{figure*}[t]
  \centering
   \includegraphics[width=0.9\linewidth]{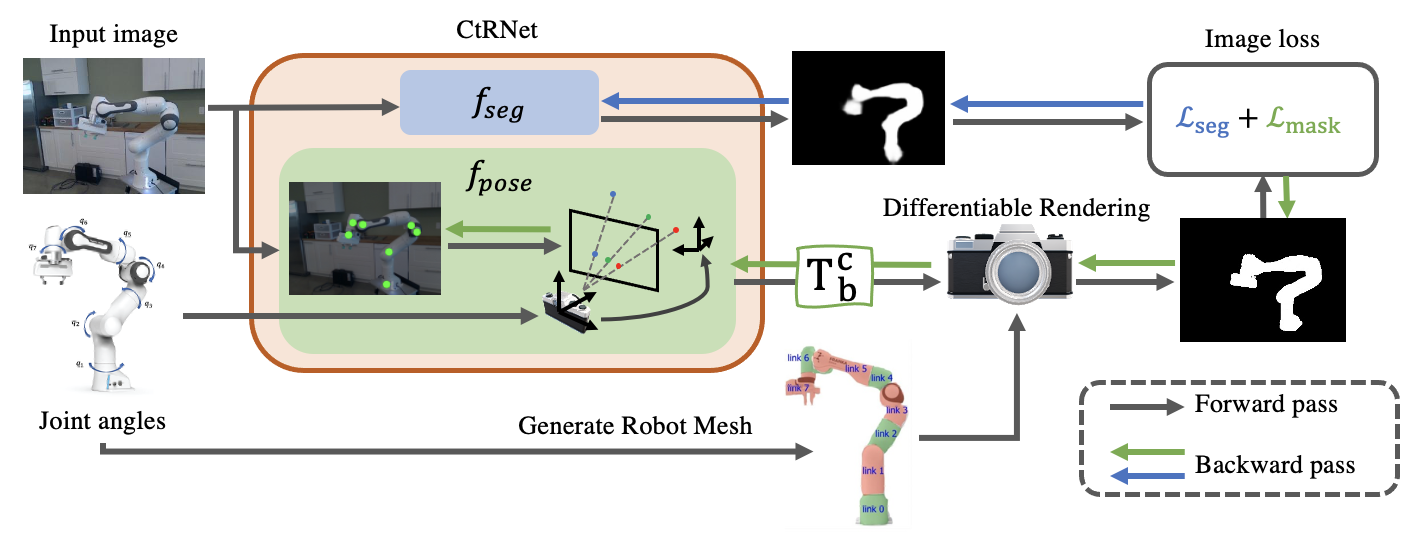}
   \caption{The overview of our proposed self-supervised training framework for sim-to-real transfer. The CtRNet contains a foreground segmentation module and a pose estimation module, which output a robot mask and a camera-to-robot pose respectively. The output pose is transformed into a silhouette image through a differentiable renderer. The image loss is back-propagated to train the keypoint detector and fine-tune the segmentation. }
   \label{fig:self_supervised_learning}
\end{figure*}

\section{Methods}
\label{sec:methods}

In this paper, we introduce an end-to-end framework for robot pose estimation and a scalable training pipeline to improve pose estimation accuracy on real-world data without the need for any manual annotation. We first explain the self-supervised training pipeline for sim-to-real transfer in \cref{subsec:sim-to-real} \textcolor{black}{given a pretrained CtRNet on synthetic data which both segments the robot and estimates its pose from images}.
Then, we detail the camera-to-robot pose estimation network in \cref{subsec:ctr_network} \textcolor{black}{which utilizes a keypoint detector and a PnP solver to estimate the pose of the robot from image data in real-time}.

\subsection{Self-supervised Training for Sim-to-Real Transfer}
\label{subsec:sim-to-real}

The most effective way to adapt the neural network to the real world is directly training the network on real sensor data.
We propose a self-supervised training pipeline for sim-to-real transfer to facilitate the training without 3D annotations.
\textcolor{black}{To conduct the self-supervised training, we employ foreground segmentation to generate a mask of the robot, $f_{seg}$, alongside the pose estimation, $f_{pose}$.}
Given an input RGB image from the physical world, $\mathbb{I}$, and the robot joint angles, $\mathbf{q}$, $f_{pose}$ estimates the robot pose which is then transformed to a silhouette image through a differentiable renderer.
Our \textcolor{black}{self-supervised} objective is to optimize neural network parameters by minimizing the difference between the rendered silhouette image and the mask image.
We formulate the optimization problem as:
\begin{multline}
    \label{eq:self_supervised_objective}
    \theta_{bb},\theta_{kp},\theta_{seg} = \argmin_{\theta_{bb},\theta_{kp},\theta_{seg} }
    \mathcal{L} [f_{seg}( \mathbb{I} |\theta_{bb},\theta_{seg}), \\ \mathcal{R}(f_{pose}(\mathbb{I} |\mathbf{q},\theta_{bb},\theta_{kp})|\mathbf{K}) ]
\end{multline}
where $\theta_{bb},\theta_{kp},\theta_{seg}$ denote the parameters of the backbone, keypoint, and segmentation layers of the neural network.
$\mathcal{R}$ is the differentiable renderer with camera parameters $\mathbf{K}$, and $\mathcal{L}(.)$ is the objective loss function capturing the image difference.

\textcolor{black}{
We pretrained CtRNet's parameters which makes up $f_{seg}$ and $f_{pose}$, with synthetic data where the keypoint and segmentation labels are obtained freely (details in Supplementary Materials).
During the self-training phase, where CtRNet learns with real data, the objective loss in (\ref{eq:self_supervised_objective}) captures the difference between the segmentation result and the rendered image.
The loss is iteratively back-propagated to, $\Theta$, where each iteration $f_{seg}$ and $f_{pose}$ take turns learning from each other to overcome the sim-to-real gap.
}

\textbf{Overview}.
The overview of the self-supervised training pipeline is shown in the \cref{fig:self_supervised_learning}.
\textcolor{black}{The segmentation module, $f_{seg}$, simply takes in a robot image and outputs its mask.}
The pose estimation module, $f_{pose}$, consists of a keypoint detector and a PnP solver to estimate the robot pose using the 2D-3D point correspondence, as shown in \cref{fig:network}. 
Given the input robot image and joint angles, our camera-to-robot pose estimation network outputs a robot mask and the robot pose with respect to the camera frame.
Mathematically, these functions are denoted as
\begin{equation}
    \mathbb{M} = f_{seg} (\mathbb{I} | \theta_{bb},\theta_{seg}) \qquad \mathbf{T}^c_b = f_{pose} (\mathbb{I} | \mathbf{q}, \theta_{bb},\theta_{kp})
\end{equation}
where $\mathbb{M}$ is the robot mask and $\mathbf{T}^c_b \in SE(3)$ is the 6-DOF robot pose.
\textcolor{black}{Finally, the self-supervised objective loss in (\ref{eq:self_supervised_objective}) is realized through a differentiable renderer, $\mathcal{R}$, which generates a silhouette image of the robot given its pose, $\mathbf{T}^{c}_b$.}

\textbf{Differentiable Rendering}.
To render the robot silhouette image, we utilize the PyTorch3D differentiable render~\cite{pytorch3d}. We initialize a perspective camera with intrinsic parameters $\mathbf{K}$ and a silhouette renderer, which does not apply any lighting nor shading, is constructed with a rasterizer and a shader.
The rasterizer applies the fast rasterization method~\cite{pytorch3d} which selects the $k$ nearest mesh triangles that effects each pixel and weights their influence according to the distance along the $z$-axis.
Finally, the \textit{SoftSilhouetteShader} is applied to compute pixel values of the rendered image using the sigmoid blending method~\cite{liu2019soft}.

We construct the ready-to-render robot mesh by connecting the CAD model for each robot body part using its forward kinematics and transforming them to the camera frame with the estimated robot pose $\mathbf{T}^c_b$ from $f_{pose}$. 
Let $\mathbf{v}^n \in \mathbb{R}^3$ be a mesh vertex on the $n$-th robot link.
Each vertex is transformed to the camera frame, hence ready-to-render, by
\begin{equation}
\label{eq:vertex_transform}
    \overline{\mathbf{v}}^c = \mathbf{T}^c_b \mathbf{T}^b_n(\mathbf{q}) \overline{\mathbf{v}}^n
\end{equation}
where $\overline{\cdot}$ represents the homogeneous representation of a point (e.g. $\overline{\mathbf{v}} = [\mathbf{v}, 1]^T$), and $\mathbf{T}^b_n(\mathbf{q})$ is the coordinate frame transformation obtained from the forward kinematics~\cite{denavit1955kinematic}.

\textbf{Objective loss function}. \textcolor{black}{The objective loss in (\ref{eq:self_supervised_objective}) is iteratively minimized where $f_{seg}$ and $f_{pose}$ take turns supervising each other on real data to overcome the sim-to-real gap faced by keypoint detection networks.}
\textcolor{black}{
To optimize $f_{pose}$, the L2 image loss is used since the segmentation network's accuracy, within the context of estimating robot poses, has been shown to effectively transfer from simulation to the real world \cite{labbe2021robopose}.
Mathematically the loss is expressed as}
\begin{equation}
        \mathcal{L}_{mask} = \sum_{i=1}^{H} \sum_{j=1}^{W} \left(\mathbb{S}(i,j) - \mathbb{M}(i,j)\right)^2
\label{eq:mask_loss}
\end{equation}
where $H$ and $W$ is the height and width of the image, and $\mathbb{S}$ is the rendered silhouette image.

Although the pretrained robot segmentation, $f_{seg}$, already performs well on real-world datasets, it is still desirable to refine it through self-supervised training to better extract fine details of corners and boundaries.
To prevent the foreground segmentation layers from receiving noisy training signals, we apply the weighted Binary Cross Entropy Loss so that the high-quality rendering image can be used to further refine the foreground segmentation:
\begin{multline}
    \mathcal{L}_{seg} = - \frac{w}{H*W} \sum_{i=1}^{H} \sum_{j=1}^{W} [\mathbb{M}(i,j) \log \mathbb{S}(i,j) \\ + (1 - \mathbb{M}(i,j)) \log(1-\mathbb{S}(i,j)) ].
\end{multline}
where $w$ is the weight for the given training sample.
For PnP solvers, the optimal solution should minimize the point reprojection error. Therefore, we assign the weight for each training sample according to the reprojection error:
\begin{equation}
    w = exp\left( - s
    O(\mathbf{o}, \mathbf{p}, \mathbf{K}, \mathbf{T}_b^c)
    \right)
    \label{eq:weights}
\end{equation}
where $s$ is a scaling constant,  $O$ is the reprojection loss in the PnP solver (explained in \cref{subsec:ctr_network}), $\{\mathbf{o}_{i} | \mathbf{o}_{i} \in \mathbb{R}^2 \}^{n}_{i=1}$ and $\{\mathbf{p}_{i} | \mathbf{p}_{i} \in \mathbb{R}^3 \}^{n}_{i=1}$ are the 2D-3D keypoints inputted into the PnP solver. 
\textcolor{black}{
The exponential function is applied to the weight such that training samples with poor PnP convergence are weighted exponentially lower than good PnP convergence thereby stabilizing the training.
}

\subsection{Camera-to-Robot Pose Estimation Network}
\label{subsec:ctr_network}

\begin{figure}[t]
  \centering
   \includegraphics[width=1.0\linewidth]{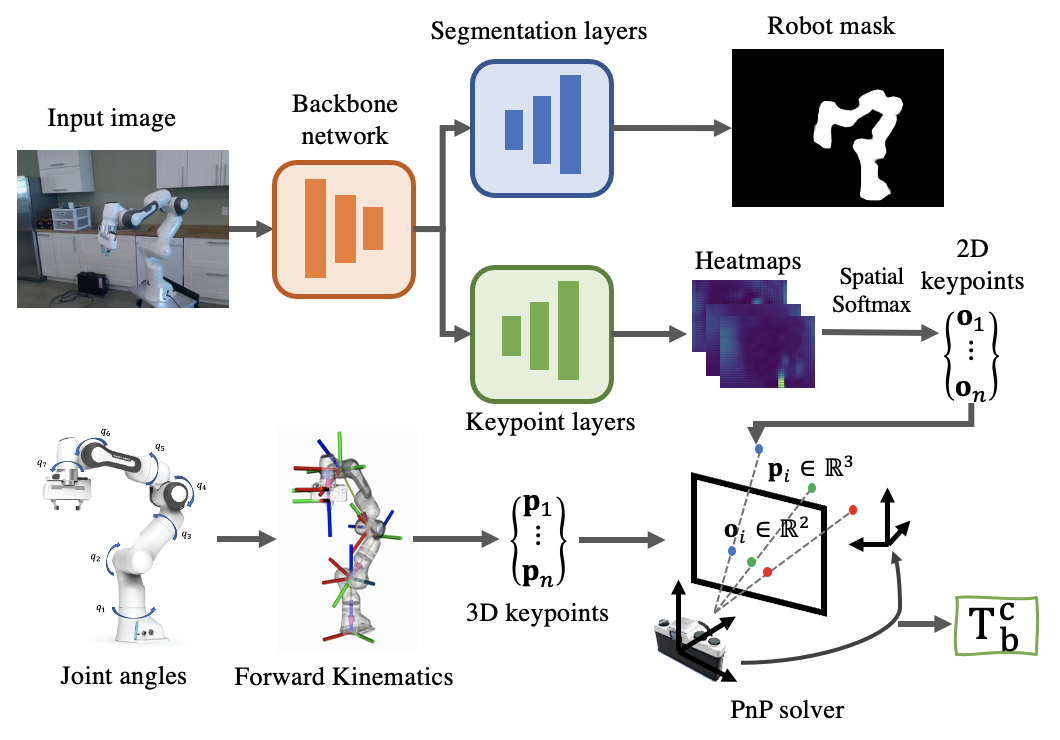}
   \caption{The diagram of the camera-to-robot pose estimation network (CtRNet) which describes the inference process of network. Given an input RGB image, the neural network generates a robot mask and a set of keypoint. Given the associated robot joint angles, a set of corresponding 3D keypoints are computed with forward kinematics. The camera-to-robot pose is estimated by a PnP solver with provided 2D-3D keypoint pairs.}
   \label{fig:network}
   \vspace{-0.18in}
\end{figure}

The overview of the proposed Camera-to-Robot Pose Estimation Network, CtRNet, is shown in Fig. \ref{fig:network}. 
Given an input RGB image, we employ ResNet50~\cite{he2016deep} as the backbone network to extract the latent features.
The latent features are then passed through the Atrous Spatial Pyramid Pooling layers~\cite{chen2017rethinking} to form the segmentation mask of input resolution.
The keypoint detector, sharing the backbone network with the foreground segmentation, upsamples the feature maps through transposed convolutional layers and forms the heatmaps with $n$ channels.
Then, we apply the spatial softmax operator~\cite{finn2016deep} on the heatmaps, which computes the expected 2D location of the points of maximal activation for each channel and results in a set of keypoints $[\mathbf{o}_1, ..., \mathbf{o}_n]$ for all $n$ channels.
For simplicity, we define the set of keypoints at each joint location of the robot.
Given the joint angles, the corresponding 3D keypoint location $\mathbf{p}_i$ can be calculated with robot forward kinematics:
\begin{equation}
    \overline{\mathbf{p}}_i = \mathbf{T}^b_i (\mathbf{q})\overline{\mathbf{t}}, \:for \: i = 1,...,n
    \label{eq:fk}
\end{equation}
where $\mathbf{t} = [0,0,0]$. With the 2D and 3D corresponding keypoints, we can then apply a PnP solver~\cite{lepetit2009epnp} to estimate the robot pose with respect to the camera frame.

\textbf{Back-propagation for PnP Solver}.
A PnP solver is usually self-contained and not differentiable as the gradient with respect to the input cannot be derived explicitly. Inspired by~\cite{bpnp}, the implicit function theorem~\cite{krantz2002IFT} is applied to obtain the gradient through implicit differentiation.
Let the PnP solver be denoted as followed in the form of a non-linear function $g$:
\begin{equation}
    \mathbf{T}_b^{c*} = g(\mathbf{o}, \mathbf{p}, \mathbf{K})
\end{equation}
where $\mathbf{T}_b^{c*}$ is output pose from the PnP solver. In order to back-propagate through the PnP solver for training the keypoint detector, we are interested in finding the gradient of the output pose $\mathbf{T}_b^{c*}$ with respect to the input 2D points $\mathbf{o}$.
Note that, the objective of the PnP solver is to minimize the reprojection error, such that:
\begin{equation}
    \mathbf{T}_b^{c*} = \argmin_{\mathbf{T}_b^{c}} O(\mathbf{o}, \mathbf{p}, \mathbf{K}, \mathbf{T}_b^{c}) \label{eq:bpnp_objective}
\end{equation}
with
\begin{align}
    O(\mathbf{T}_b^{c}, \mathbf{p}, \mathbf{K}, \mathbf{T}_b^{c}) & =
     \sum_{i=1}^n || \mathbf{o}_i - \pi(\mathbf{p}_i| \mathbf{T}_b^{c}, \mathbf{K}) ||^2_2 \label{eq:bpnp_r}\\
    & =  \sum_{i=1}^n || \mathbf{r}_i ||_2^2 \label{eq:bpnp_r_2}
\end{align}
where $\pi(.)$ is the projection operator. Since the optimal solution $\mathbf{T}_b^{c*}$ is a local minimum for the objective function $O(\mathbf{o}, \mathbf{p}, \mathbf{T}_b^{c}, \mathbf{K})$, a stationary constraint of the optimization process can be constructed by taking the first order derivative of the objective function with respect to $\mathbf{T}_b^{c}$:
\begin{equation}
    \frac{\partial O}{\partial \mathbf{T}_b^{c}} (\mathbf{o}, \mathbf{p}, \mathbf{K}, \mathbf{T}_b^{c})|_{\mathbf{T}_b^{c} = \mathbf{T}_b^{c*}} = \mathbf{0}.
\end{equation}
Following \cite{bpnp}, we construct a constrain function $F$ to employ the implicit function theorem:
\begin{equation}
    F(\mathbf{o}, \mathbf{p}, \mathbf{K}, \mathbf{T}_b^{c}) = \frac{\partial O}{\partial \mathbf{T}_b^{c}} (\mathbf{o}, \mathbf{p}, \mathbf{K}, \mathbf{T}_b^{c*}) = \mathbf{0}.
    \label{eq:bpnp_constraint_function}
\end{equation}
Substituting the \cref{eq:bpnp_r} and \cref{eq:bpnp_r_2} to \cref{eq:bpnp_constraint_function}, we can derive the constraint function as:
\begin{align}
    F(\mathbf{o}, \mathbf{p}, \mathbf{K}, \mathbf{T}_b^{c}) 
    & = \sum_{i=1}^n \frac{\partial || \mathbf{r}_i ||_2^2}{\partial \mathbf{T}_b^{c}} \\
    & = -2 \sum_{i=1}^n \mathbf{r}_i^T \frac{\partial \pi}{\partial \mathbf{T}_b^{c}}(\mathbf{p}_i| \mathbf{T}_b^{c*}, \mathbf{K}) .
\end{align}
Finally, we back-propagate through the PnP solver with the implicit differentiation. The gradient of the output pose with respect to the input 2D points is the Jacobian matrix:
\begin{multline}
    \frac{\partial g}{\partial \mathbf{o}}(\mathbf{o}, \mathbf{p}, \mathbf{K}) \\ = - \left( \frac{\partial F}{\partial \mathbf{T}_b^{c}} (\mathbf{o}, \mathbf{p}, \mathbf{K}, \mathbf{T}_b^{c}) \right)^{-1}\left( \frac{\partial F}{\partial \mathbf{o}} (\mathbf{o}, \mathbf{p}, \mathbf{K}, \mathbf{T}_b^{c})\right) .
\end{multline}

\section{Experiments}
\label{sec:experiments}

We first evaluate our method on two public real-world datasets for robot pose estimation and compare it against several state-of-the-art image-based robot pose estimation algorithms. 
We then conduct an ablation study on the pretraining procedure and explore how the number of pretraining samples could affect the performance of the self-supervised training.
Finally, we integrate the camera-to-robot pose estimation framework into a visual servoing system to demonstrate the effectiveness of our method on real robot applications.

\begin{table*}
\setlength\tabcolsep{0.34em}
  \centering
  \scalebox{0.8}{
  \begin{tabular}{@{}lccccccccccc c@{}}
    \toprule
    \multirow{2}{*}{Method} & \multirow{2}{*}{Category} & \multirow{2}{*}{Backbone} & \multicolumn{2}{c}{Panda 3CAM-AK}  &  \multicolumn{2}{c}{Panda 3CAM-XK} & \multicolumn{2}{c}{Panda 3CAM-RS} &  \multicolumn{2}{c}{Panda ORB}&  \multicolumn{2}{c}{All} \\
    \cmidrule(lr){4-5} \cmidrule(lr){6-7} \cmidrule(lr){8-9} \cmidrule(lr){10-11} \cmidrule(lr){12-13}
     &  &  & AUC $\color{green}\uparrow$ & Mean (m) $\color{red}\downarrow$ 
     & AUC $\color{green}\uparrow$ & Mean (m) $\color{red}\downarrow$ 
     & AUC $\color{green}\uparrow$ & Mean (m) $\color{red}\downarrow$ 
     & AUC $\color{green}\uparrow$ & Mean (m) $\color{red}\downarrow$ 
     & AUC $\color{green}\uparrow$ & Mean (m) $\color{red}\downarrow$ \\
    \midrule
    DREAM-F~\cite{lee2020dream} & Keypoint & VGG19 & 68.912 & 11.413 & 24.359 & 491.911 & 76.130 & 2.077 & 61.930 & 95.319 & 60.740 & 113.029\\
    DREAM-Q~\cite{lee2020dream} & Keypoint & VGG19 & 52.382 & 78.089 & 37.471 & 54.178 & 77.984 & 0.027 & 57.087 & 67.248 & 56.988 & 59.284\\
    DREAM-H~\cite{lee2020dream} & Keypoint & ResNet101 & 60.520 & 0.056 & 64.005 & 7.382 & 78.825 & 0.024 & 69.054 & 25.685 & 68.584 & 17.477\\
    RoboPose~\cite{labbe2021robopose} & Rendering & ResNet34 & 76.497  & 0.024  & \textbf{85.926} & \textbf{0.014} & 76.863 & 0.023 & 80.504 & \textbf{0.019} & 80.094 & \textbf{0.020}\\
    CtRNet &  Keypoint &   ResNet50 & \textbf{89.928} & \textbf{0.013} &   79.465 & 0.032 & \textbf{90.789} & \textbf{0.010} & \textbf{85.289} & 0.021 & \textbf{85.962} & \textbf{0.020}\\
    \bottomrule
  \end{tabular}}
  \caption{Comparison of our methods with the state-of-the-art methods on DREAM-real datasets using ADD metric. We report the mean and AUC of the ADD on each dataset and the overall accuracy. }
  \label{table:deam}
\end{table*}

\begin{table*}[t]
\centering
\setlength\tabcolsep{0.34em}
\scalebox{0.95}{
\begin{tabular}{lccccccc}
\toprule
\multirow{2}{*}{Method} & \multirow{2}{*}{Category}  & \multicolumn{2}{c}{PCK (2D)} & Mean 2D Err. & \multicolumn{2}{c}{ADD (3D)}& Mean 3D Err.\\ 
\cmidrule(lr){3-4} \cmidrule(lr){6-7}
  & & @50 pixel $\color{green}\uparrow$ & AUC $\color{green}\uparrow$ & (pixel) $\color{red}\downarrow$ & @100 mm $\color{green}\uparrow$ & AUC $\color{green}\uparrow$ & (mm) $\color{red}\downarrow$ \\ \midrule
Aruco Marker~\cite{garrido2014aruco}    & Keypoint & 0.49  & 57.15 & 286.98 & 0.30 & 43.45 & 2447.34\\
DREAM-Q~\cite{lee2020dream} & Keypoint & 0.33 & 44.01 & 1277.33 & 0.32  & 40.63 & 386.17 \\
Opt. Keypoints~\cite{lu2022keypoint}  & Keypoint & 0.69  & 75.46 & 49.51 & 0.47  &65.66 & 141.05 \\
Diff. Rendering & Rendering& 0.74  & 78.60 & 42.30  & 0.78  & 81.15 & 74.95\\
CtRNet            & Keypoint & \textbf{0.99}  & \textbf{93.94} & \textbf{11.62} & \textbf{0.88} & \textbf{83.93} & \textbf{63.81}\\
\bottomrule
\end{tabular}}
\caption{\label{exp:rigid_pck}Comparison of our methods with the state-of-the-art methods on Baxter dataset.}
\label{table:baxter}
\end{table*}

\subsection{Datasets and Evaluation Metrics}

\textbf{DREAM-real Dataset}. The DREAM-real dataset~\cite{lee2020dream} is a real-world robot dataset collected with 3 different cameras: Azure Kinect (AK), XBOX 360 Kinect (XK), and RealSense (RS). This dataset contains around 50K RGB images of Franka Emika Panda arm and is recorded at ($640 \times 480$) resolution. The ground-truth camera-to-robot pose is provided for every image frame.
The accuracy is evaluated with average distance (ADD) metric~\cite{xiang2017posecnn},
\begin{equation}
    ADD = \frac{1}{n} \sum_{i=1}^n || \widetilde{\mathbf{T}}^c_b 
    \overline{\mathbf{p}}_i - \mathbf{T}^c_b 
    \overline{\mathbf{p}}_i ||_2
\end{equation}
where $\widetilde{\mathbf{T}}^c_b $ indicates the ground-truth camera-to-robot pose.
We also report the area-under-the-curve (AUC) value, which integrates the percentage of ADD over different thresholds. A higher AUC value indicates more predictions with less error.

\textbf{Baxter Dataset}. The Baxter dataset~\cite{lu2022keypoint} contains 100 RGB images of the left arm of Rethink Baxter collected with Azure Kinect camera at ($2048 \times1526$) resolution. The 2D and 3D ground-truth end-effector position with respect to the camera frame is provided. 
We evaluate the performance with the ADD metric for the end-effector.
We also evaluate the end-effector reprojection error using the percentage of correct keypoints (PCK) metric~\cite{lu2022keypoint}.

\begin{figure*}[t]
\centering
\subfloat[]{\includegraphics[width=0.48\linewidth]{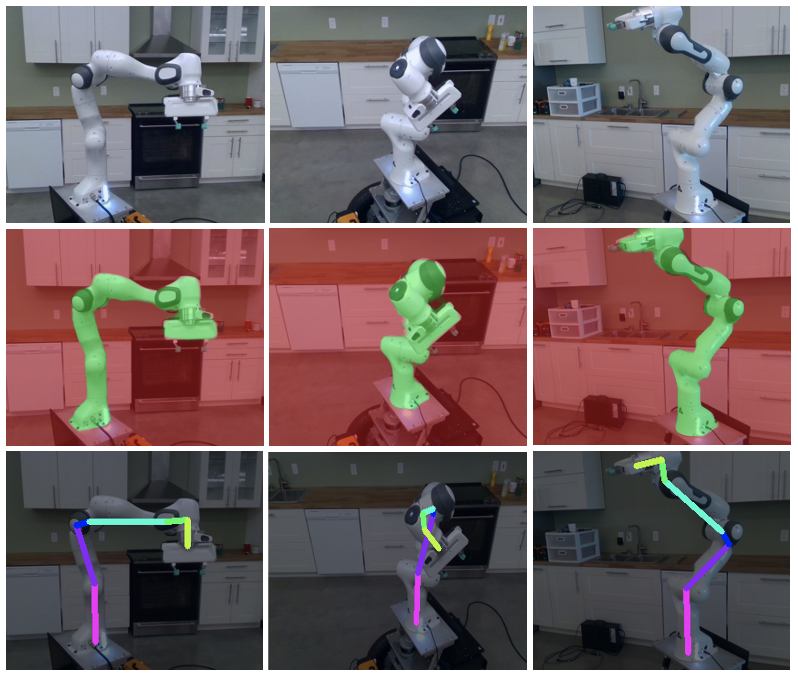}\label{fig:panda}}
\subfloat[]{\includegraphics[width=0.48\linewidth]{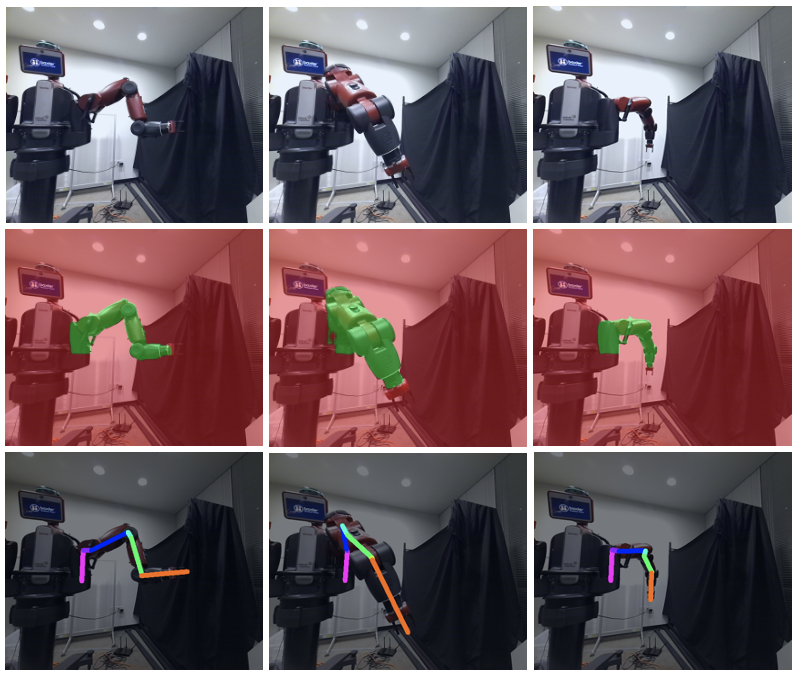}\label{fig:baxter}}
\caption{Qualitative results of CtRNet foreground segmentation and pose estimation on \protect\subref{fig:panda}~DREAM-real dataset and \protect\subref{fig:baxter}~Baxter dataset. The first row shows the input RGB image, the second row shows the foreground segmentation, and the third row shows the projected robot skeleton based on the estimated robot pose.}
\label{fig:qualitative_results}
\end{figure*}

\subsection{Implementation details}
The entire pipeline is implemented in PyTorch~\cite{torchautograd}. We initialize the backbone network with ImageNet~\cite{deng2009imagenet} pretrained weights, and we train separate networks for different robots.
The number of keypoints $n$ is set to the number of robot links and the keypoints are defined at the robot joint locations.
The neural network is pretrained on synthetic data for foreground segmentation and keypoint detection for 1000 epochs with 1e-5 learning rate. We reduce the learning rate by a factor of 10 once learning stagnates for 5 epochs. The Adam optimizer is applied to optimize the network parameters with the momentum set to 0.9.
For self-supervised training on real-world data, we run the training for 500 epochs with 1e-6 learning rate. The same learning rate decay strategy and Adam optimizer is applied here similar to the pretraining. To make the training more stable, we clip the gradient of the network parameters at 10. The scaling factor in \cref{eq:weights} is set to 0.1 for DREAM-real dataset and 0.01 for Baxter dataset, mainly accounting for the difference in resolution.

\subsection{Robot Pose Estimation on Real-world Datasets}

\textbf{Evaluation on DREAM-real Dataset}. The proposed CtRNet is trained at ($320 \times 240$) resolution and evaluated at the original resolution by scaling up the keypoints by a factor of 2.
Some qualitative results for foreground segmentation and pose estimation are shown in the \cref{fig:panda}.
We compared our method with the state-of-the-art keypoint-based method DREAM~\cite{lee2020dream} and the rendering-based method RoboPose~\cite{labbe2021robopose}. 
The results for DREAM and RoboPose are compiled from the implementation provided by~\cite{labbe2021robopose}.
In \cref{table:deam}, we report the AUC and mean ADD results on DREAM-real dataset with 3 different camera settings and the overall results combining all the test samples.
Our method has a significantly better performance compared to the method in the same category and achieves comparable performance with the rendering-based method.
We outperform DREAM on all settings and outperform RoboPose on the majority of the dataset. 
Overall on DREAM-real dataset, we achieve higher AUC (+17.378 compared to DREAM, +5.868 compared to RoboPose), and lower error compared to DREAM (-17.457).

\textbf{Evaluation on Baxter Dataset}. For the Baxter dataset, we trained the CtRNet at ($640\times480$) resolution and evaluate at the original resolution, and \cref{fig:baxter} shows some of the qualitative results.
We compared our method with several keypoint-based methods (Aruco Marker~\cite{garrido2014aruco}, DREAM~\cite{lee2020dream}, Optimized Keypoints~\cite{lu2022keypoint}).
We also implemented Differentiable Rendering for robot pose estimation, where the robot masks are generated with the pretrained foreground segmentation.
The 2D PCK results and 3D ADD results are reported in \cref{table:baxter}.
Our method outperforms all other methods on both 2D and 3D evaluations.
For 2D evaluation, we achieve 93.94 AUC for PCK with an average reprojection error of 11.62 pixels.
For 3D evaluation, we achieve 83.93 AUC for ADD with an average ADD of 63.81mm.
Notably, 99 percent of our estimation has less than 50 pixel reprojection error, which is less than 2 percent of the image resolution, and 88 percent of our estimation has less than 100mm distance error when localizing the end-effector.

\subsection{Ablation Study}

\begin{figure}[t]
\centering
\input{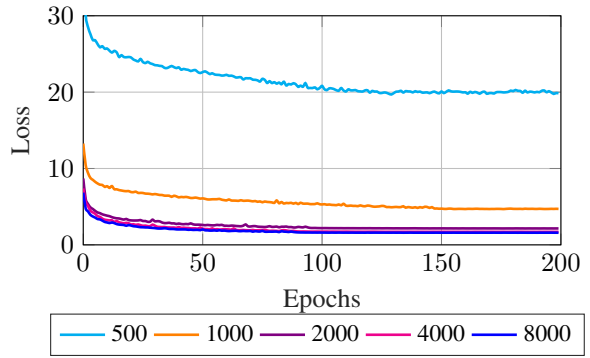}
\caption{\label{fig:abalation} The training loss vs. number of epochs for the self-supervised training with different numbers of pretraining samples. More pretraining samples results in better convergence. The improvement saturates after having more than 2000 pretraining samples as only marginal improvement by adding more samples.}
\end{figure}

We study how the number of pretraining samples affects the convergence and performance of the self-supervised training empirically on the Baxter dataset. 
We pretrain the neural network with different numbers of synthetic data samples $N_{pretrain} = \{500,1000,2000,4000,8000\}$, and examine the convergence of the self-supervised training process.
\cref{fig:abalation} shows the plot of self-training loss ($\mathcal{L}_{mask}+ \mathcal{L}_{seg}$) vs. the number of epochs for networks pretrianed with different number of synthetic data. We observe that doubling the size of pretraining dataset significantly improves the convergence of the self-training process at the beginning. However, the improvement gets smaller as the pretrainig size increase. For the Baxter dataset, the improvement saturates after having more than 2000 pretraining samples. Continuing double the training size results in very marginal improvement.
Noted that the Baxter dataset captures 20 different robot poses from a fixed camera position. The required number of pretraining samples might vary according to the complexity of the environment. 

We further evaluate the resulting neural networks with the ground-truth labels on the Baxter dataset. We report the mean ADD and AUC ADD for the pose estimation in \cref{tab:ablation}. The result verifies our observation on the convergence analysis. 
Having more pretraining samples improves the performance of pose estimation at the beginning, but the improvement stagnates after having more than 2000 pretraining samples.

\begin{table}
  \centering
  \begin{tabular}{@{}lcc@{}}
    \toprule
    $N_{pretrain}$ & Mean ADD (mm) $\color{red}\downarrow$ & AUC ADD $\color{green}\uparrow$\\
    \midrule
    500 & 2167.30 & 47.62 \\
    1000 & 92.91 & 76.65 \\
    2000 & 67.51 & 82.98 \\
    4000 & 63.00 & 84.12 \\
    8000 & 63.81 & 83.93\\
    \bottomrule
  \end{tabular}
  \caption{Ablation study for the number of pretraining samples.}
  \label{tab:ablation}
\end{table}

\subsection{Visual Servoing Experiment}

\begin{table}
  \centering
  \scalebox{0.9}{
  \begin{tabular}{@{}lccc@{}}
    \toprule
    Method & Loop Rate & Trans. Err. (m) & Rot. Err. (rad)\\
    \midrule
    DREAM~\cite{lee2020dream} & 30Hz & 0.235 $\pm$ 0.313 & 0.300 $\pm$ 0.544 \\
    Diff. Rendering & 1Hz & 0.046 $\pm$ 0.062 & 0.036 $\pm$ 0.066\\
    CtRNet & 30Hz & \textbf{0.002 $\pm$ 0.001} & \textbf{0.002 $\pm$ 0.001}\\
    \bottomrule
  \end{tabular}}
  \caption{Mean and standard deviation of the translational error and rotational error for the visual servoing experiment.}
  \label{tab:visual_servoing}
\end{table}

We integrate the proposed CtRNet into a robotic system for position-based visual servoing (PBVS) with eye-to-hand configuration. 
We conduct the experiment on a Baxter robot and the details of the PBVS are described in the Supplementary Materials.
The PBVS is purely based on RGB images from a single camera and the goal is to control the robot end-effector reaching a target pose defined in the camera frame.
Specifically, we first set a target pose with respect to the camera frame. The target pose is then transformed into the robot base frame through the estimated camera-to-robot transformation. The robot controller calculates the desired robot configuration with inverse kinematics and a control law is applied to move the robot end-effector toward the target pose.

\begin{figure}[t]
\centering
\includegraphics[width=0.95\linewidth]{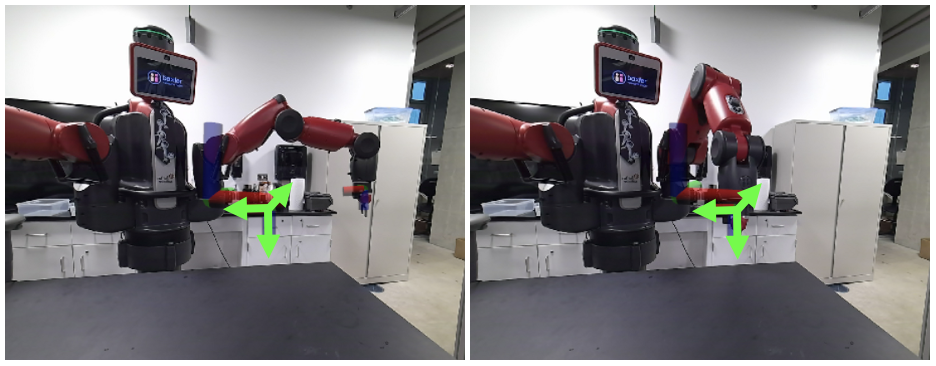}
\caption{Snapshots of PBVS. The goal is to move the end-effector to the target pose (green). The figure on the right shows the robot configuration upon the convergence of PBVS.}
\label{fig:snapshot}
\end{figure}

For comparison, we also implemented DREAM~\cite{lee2020dream} and a Differentiable Renderer for PBVS. For DREAM, the pretrained model for Baxter is applied. For Differentiable Renderer, we use the foreground segmentation of CtRNet to generate a robot mask. The optimizing loop for the renderer takes the last estimation as initialization and performs 10 updates at each callback to ensure convergence and maintain 1Hz loop rate.
In the experiment, we randomly set the target pose and the position of the camera, and the robotic system applies PBVS to reach the target pose from an arbitrary initialization, as shown in the \cref{fig:snapshot}.  
We ran the experiment for 10 trails with different robot pose estimation methods, and the translational (Euclidean distance) and rotational errors (Euler angles) of the end-effector are reported in \cref{tab:visual_servoing}.
The experimental results show that our proposed method significantly improves the stability and accuracy of the PBVS, achieving 0.002m averaged translational error and 0.002rad rotational error on the end-effector.

\begin{figure}[t]
\centering
\input{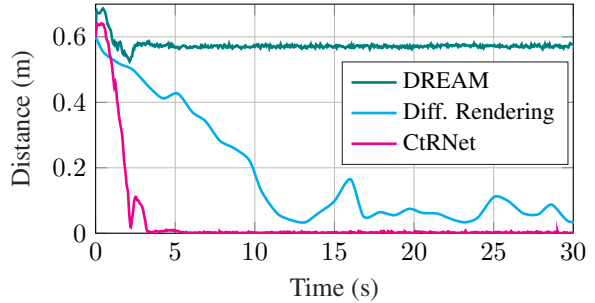}
\caption{The plot of end-effector distance-to-goal over time on a selected PBVS trail.}
\label{fig:visual_servoing}
\vspace{-0.18in}
\end{figure}

We also plot the end-effector distance-to-goal over time for a selected trail in \cref{fig:visual_servoing}.
In this selected trial, the system could not converge with DREAM because the poor robot pose estimation confuses the controller by giving the wrong target pose in the robot base frame, which is unreachable. With the differentiable renderer, the servoing system takes more than 10 seconds to converge and oscillate due to the low loop rate. With our proposed CtRNet, the servoing system converges much faster ($\leq$ 5 seconds), thanks to the fast and robust robot pose estimation. We show more qualitative results in the Supplementary Materials.

\section{Conclusion}
\label{sec:conclusion}

We present the CtRNet, an end-to-end image-based robot pose estimation framework, and a self-supervised training pipeline that utilizes unlabelled real-world data for sim-to-real transfer.
The CtRNet, using a keypoint detector for pose estimation while employing a rendering method for training, achieves state-of-the-art performance on robot pose estimation while maintaining high-speed inference. The \cref{fig:methods_scatter} illustrates the advantages of CtRNet over existing methods, where the AUC values are normalized across two evaluation datasets by taking DREAM and CtRNet as references.
We further experiment with different robot pose estimation methods by applying them to PBVS, which demonstrates CtRNet's fast and accurate robot pose estimation enabling stability when using single-frame robot pose estimation for feedback. Therefore, CtRNet supports real-time markerless camera-to-robot pose estimation which has been utilized for surgical robotic manipulation~\cite{richter2021robotic} and mobile robot manipulators~\cite{wise2016fetch}.
For future work, we would like to extend our method to more robots and explore vision-based control in an unstructured environment.

{\small
\bibliographystyle{ieee_fullname}
\bibliography{egbib}
}

\clearpage

\section{Supplementary Materials}

\subsection{Generate Synthetic Training Data}

Setting up a pipeline for generating robot masks and keypoints can be complicated or simple depending on the choice of the simulator. However, an interface for acquiring the robot pose with respect to a camera frame must exist for every robot simulator. Therefore, instead of generating ground-truth labels for the robot mask and keypoints directly from a simulator, we only save the robot pose and configuration pair for each synthetic image, and the labels for the robot mask and keypoint are generated on-the-fly during the training process.

Given the ground-truth robot pose with respect to the camera frame $\widetilde{\mathbf{T}}^c_b$ and robot configuration $\mathbf{q}$, the keypoint labels $\widetilde{\mathbf{o}}_i$ can be generated through the projection operation:
\begin{equation}
    \widetilde{\mathbf{o}}_i = \pi(\mathbf{p}_i| \widetilde{\mathbf{T}}^c_b, \mathbf{K})
\end{equation}
where $\mathbf{K}$ is the camera intrinsic matrix and $\mathbf{p}_i$ is calculated using robot forward kinematics with known robot joint angles $\mathbf{q}$ as \cref{eq:fk}.
For generating the robot mask, we apply a silhouette renderer $\mathcal{R}$ which is described in Section 3.1. Given the ground-truth robot pose, the robot mask is generated as:
\begin{equation}
    \widetilde{\mathbb{S}} = \mathcal{R}(\widetilde{\mathbf{T}}^c_b|\mathbf{K})
\end{equation}
where $\widetilde{\mathbb{S}}$ is the ground-truth label for robot mask.

For generating the synthetic images, similar to Domain Randomization~\cite{tobin2017domain}, we also randomize a few settings of the robot simulator so that the generated samples can have some varieties. Specifically, we randomize the following aspects:
\begin{itemize}
    \item Robot joint configuration.
    \item The camera position, which applies the look-at method so that the robot is always in the viewing frustum.
    \item The number, position, and intensity of the scene lights.
    \item The position of the virtual objects.
    \item The background of the images.
    \item The color of robot mesh.
\end{itemize}
We also perform image augmentation with additive white Gaussian noise.
The synthetic data for the Baxter is generated using CoppeliaSim\footnote{https://www.coppeliarobotics.com/}. For the Panda, the synthetic dataset provided from~\cite{lee2020dream} is used for training.

\begin{figure*}[t]
  \centering
   \includegraphics[width=0.95\linewidth]{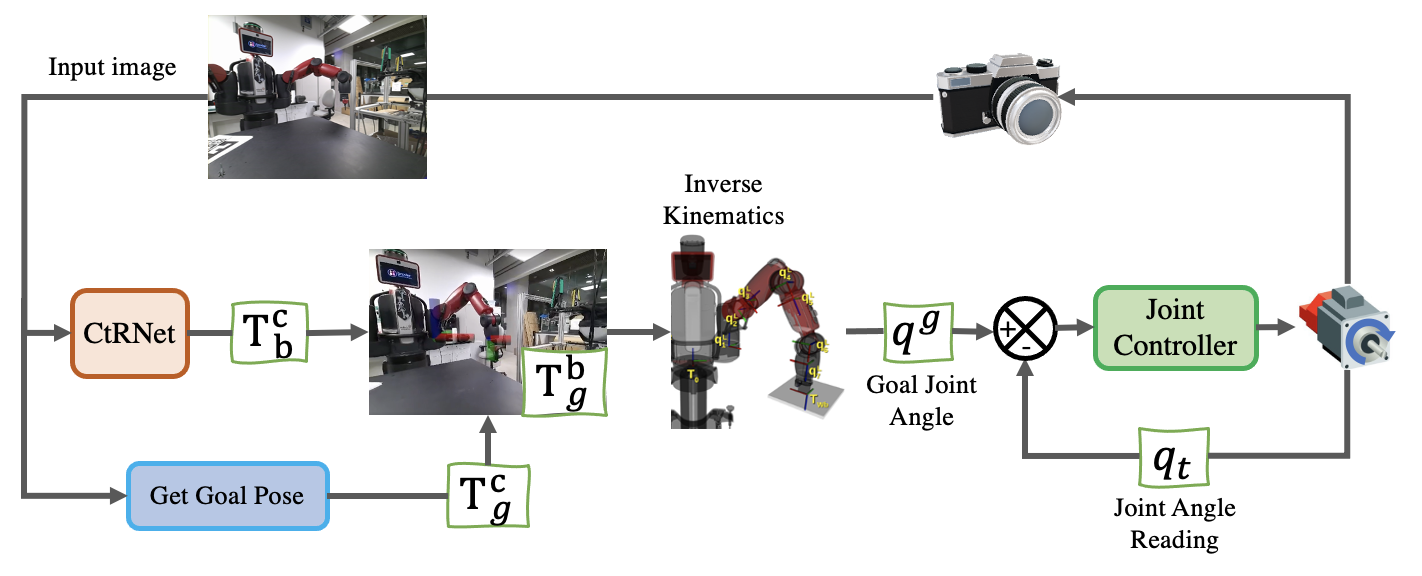}
   \caption{The diagram of the position-based visual servoing system. Given the current joint angle reading and image, CtRNet outputs the camera-to-robot pose. After transforming the goal end-effector pose to the robot base frame, the goal joint configuration is computed via inverse kinematics. Then, a joint controller is employed to minimize the error between the current and goal joint configuration.}
   \label{fig:visual_servoing_diagram}
\end{figure*}

\subsection{Details on Position-based Visual Servoing}

The diagram of the position-based visual servoing system is shown in the~\cref{fig:visual_servoing_diagram}. In our experimental setup, the camera is not necessarily stationary and the goal pose, $\mathbf{T}^c_g$, defined in the camera frame, is also changing over time. Therefore, for each control loop, we need to compute the camera-to-robot pose $\mathbf{T}^c_b$ to update the goal pose in the robot base frame:
\begin{equation}
    \mathbf{T}^b_g = [\mathbf{T}^c_b]^{-1} \mathbf{T}^c_g.
\end{equation}

Given the current joint angle reading and image, CtRNet outputs the camera-to-robot pose. The goal pose is a predefined trajectory in the camera frame. For each loop, we take the camera-to-robot pose estimation to transform the goal end-effector pose from the camera frame to the robot base frame, and the goal joint configuration is computed via inverse kinematics. Then, a joint controller is employed to minimize the joint configuration error by taking a step toward the goal joint configuration.
The camera was running at 30Hz and the joint controller was running at 120Hz. All of the communication between sub-modules was done using the Robot Operating System~(ROS), and everything ran on a single computer with an Intel\textregistered{} Core\texttrademark{} i9 Processor and NVIDIA's GeForce RTX 3090.

For qualitative analysis, we provide 2 experiment setups. At first, we fixed the goal pose at the camera center with a depth of one meter. In the second experiment, we set the goal pose following a circle centered at $[0,0,1]$, with a radius of 0.15 meters, in the camera frame.
For both experiments, we manually moved the camera, and the results are presented in the supplementary videos.

\subsection{Evaluation on Foreground Segmentation}
\begin{figure*}[t]
\centering

\includegraphics[width=0.95\linewidth]{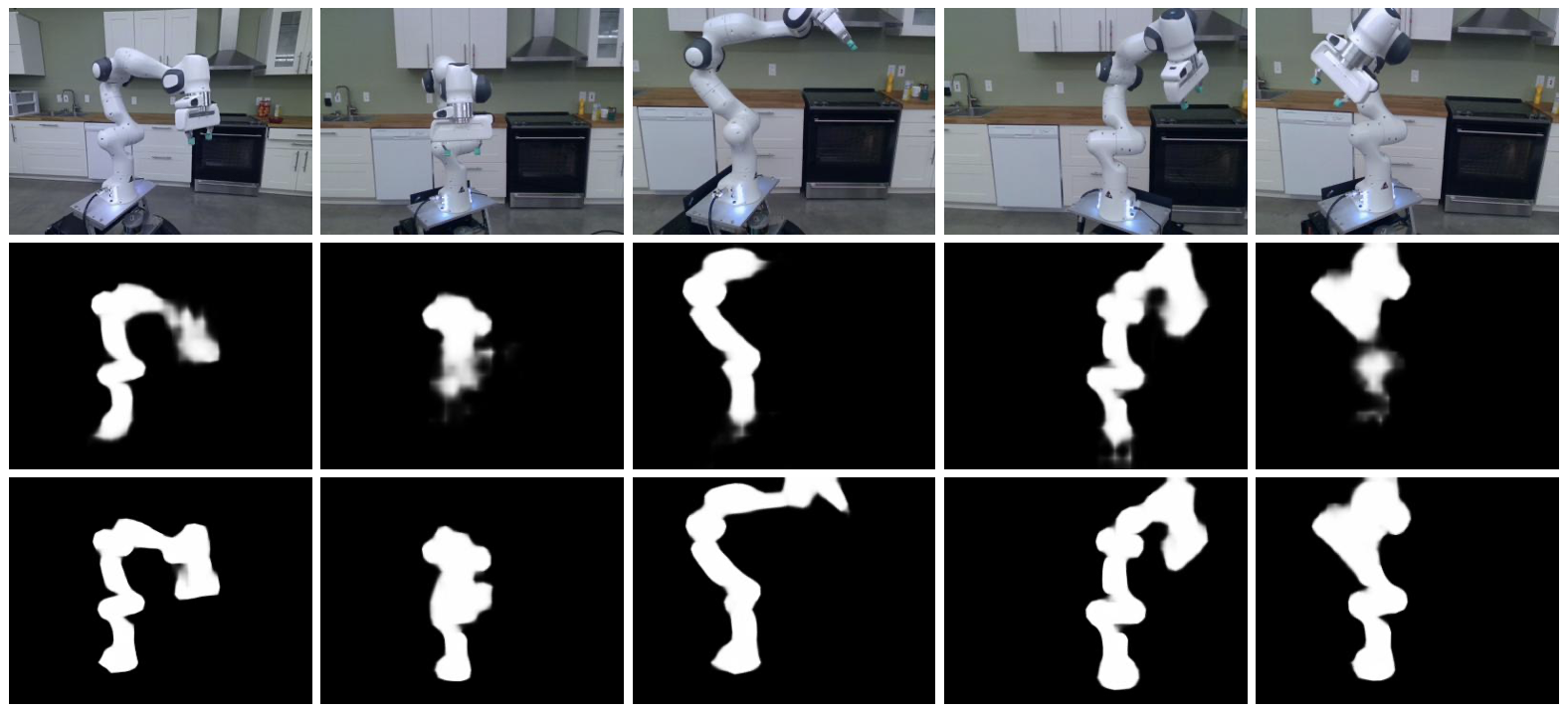}
\caption{Qualitative results of CtRNet foreground segmentation before and after self-supervised training. From top to bottom row shows the RGB images, the segmentation masks before, and after self-supervised training.}
\label{fig:segmentation_masks}
\end{figure*}

In this section, we evaluate the performance of foreground segmentation before and after the self-supervised training of the CtRNet.
As the real-world datasets do not have segmentation labels, we evaluate the segmentation performance qualitatively on the DREAM-real dataset.
The qualitative results are shown in \cref{fig:segmentation_masks}, where we show sample segmentation masks before and after the self-supervised training, together with the original RGB images. Notably, the robot mask exhibits enhanced quality after self-supervised training, preserving fine details of corners and boundaries.
We believe the high-quality segmentation mask is the key to achieving SOTA performance on robot pose estimation, which provides close-to-ground-truth supervision for the keypoint detector.

\subsection{Segmentation Accuracy Impact on Pose Estimation}

\begin{table}
  \centering
  \begin{tabular}{@{}lcccc@{}}
    \toprule
    \multirow{2}{*}{$N_{pretrain}$} & \multicolumn{2}{c}{Mean ADD (mm) $\color{red}\downarrow$} & \multicolumn{2}{c}{AUC ADD $\color{green}\uparrow$}\\ 
\cmidrule(lr){2-3} \cmidrule(lr){4-5}
& w/ $\mathcal{L}_{seg}$ & w/o $\mathcal{L}_{seg}$ & w/ $\mathcal{L}_{seg}$ & w/o $\mathcal{L}_{seg}$\\
    \midrule
    500 & 2167.30 & 184.45 & 47.62 & 65.26\\
    1000 & 92.91 & 172.34 & 76.65 & 69.71\\
    2000 & 67.51 & 104.69 & 82.98 & 76.20\\
    4000 & 63.00 & 84.67 & 84.12 & 78.71\\
    8000 & 63.81 & 83.21 & 83.93 & 79.07\\
    \bottomrule
  \end{tabular}
  \caption{Ablation study on the impact of segmentation loss.}
  \label{tab:no_seg_loss}
\end{table}

The performance of the robot pose estimation relies on the accuracy of the foreground segmentation, as the segmentation mask offers image-level guidance during the self-supervised training phase. In this section, we investigate the influence of segmentation accuracy on the performance of robot pose estimation.
To analyze this, we carry out experiments using the Baxter dataset by training the neural network with and without the segmentation loss $\mathcal{L}_{seg}$. We use different numbers of synthetic training samples to pre-train the neural network as described in Section 4.4. 
Throughout the self-supervised training phase, only the mask loss $\mathcal{L}{mask}$ is utilized to train the keypoint detector, while refraining from fine-tuning the segmentation layers with segmentation loss $\mathcal{L}_{seg}$.

We report the mean ADD and AUC ADD for robot pose estimation in~\cref{tab:no_seg_loss}, and include the results obtained with segmentation loss $\mathcal{L}_{seg}$ for comparative purposes. 
We observed that training with segmentation loss consistently yields better performance by a considerable margin, given enough pertrained samples. We also discovered a similar trend as in Section 4.4: having more pretraining samples improves the performance but the improvement is saturated after having more than 4000 training samples. However, when limited to a low number of pretraining samples ($\leq 500$), CtRNet performs better without using segmentation loss because the segmentation layers are not affected by the large numbers of inaccurately detected keypoints.

\subsection{Limitation}
The keypiont detection can only detect the points in the image frame. In some cases, part of the robot body is out of the camera frustum and hence does not appear in the image frame.
This will result in false positives in keypoint detection, which undermines the performance of the robot pose estimation.
However, the false positives in keypoint usually result in a pose that has a large reprojection error. Therefore, the foreground segmentation would not affect by these bad samples as the weights are close to zero, according to \cref{eq:weights}. During the inference, we can also use the reprojection error to indicate the confidence of the pose estimation.

\end{document}